\documentclass{IEEEtran}
\usepackage[style=ieee, maxnames=3, minnames=1, backend=biber]{biblatex}
\usepackage{enumitem}
\usepackage{amsmath,amssymb,amsfonts}
\usepackage{algorithmic}
\usepackage{graphicx}
\usepackage{textcomp}
\usepackage{array}
\usepackage{tabularx}
\usepackage{lipsum} 
\usepackage{multirow}
\usepackage{makecell}
\usepackage{color}

\def\BibTeX{{\rm B\kern-.05em{\sc i\kern-.025em b}\kern-.08em
    T\kern-.1667em\lower.7ex\hbox{E}\kern-.125emX}}
\begin{document}

\title{Artificial General Intelligence\\for Medical Imaging Analysis}
\author{Xiang Li*, Lin Zhao*, Lu Zhang*, Zihao Wu, Zhengliang Liu, Hanqi Jiang, Chao Cao, Shaochen Xu, Yiwei Li, Haixing Dai, Yixuan Yuan, Jun Liu, Gang Li, Dajiang Zhu, Pingkun Yan, Quanzheng Li, Wei Liu, Tianming Liu~\IEEEmembership{Senior Member, IEEE}, and Dinggang Shen~\IEEEmembership{Fellow, IEEE}
\thanks{*These authors contributed equally to this work.}
\thanks{\textit{(Corresponding authors: Xiang Li, Tianming Liu, Dinggang Shen)}}
\thanks{Xiang Li and Quanzheng Li are with the Department of Radiology, Massachusetts General Hospital and Harvard Medical School, Boston 02115, USA. (e-mail: \{xli60,li.quanzheng\}@mgh.harvard.edu).}
\thanks{Lu Zhang, Chao Cao, and Dajiang Zhu are with the Department of Computer Science and Engineering, The University of Texas at Arlington, Arlington 76019, USA. (e-mail: lu.zhang2@mavs.uta.edu, cxc0366@mavs.uta.edu, and dajiang.zhu@uta.edu).}
\thanks{Lin Zhao, Zihao Wu, Zhengliang Liu, Shaochen Xu, Yiwei Li, Haixing Dai and Tianming Liu are with the School of Computing, The University of Georgia, Athens 30602, USA. (e-mail: \{zihao.wu1,zl18864,lin.zhao, hd54134, tliu\}@uga.edu).}
\thanks{Hanqi Jiang is with the School of Computer and Information Technology, Beijing Jiaotong University, Beijing 100044, China; College of Engineering, The University of Georgia, Athens 30602, USA. (e-mail: hj67104@uga.edu).}
\thanks{Yixuan Yuan is with the Department of Electronic Engineering, Chinese University of Hong Kong, Hong Kong. (e-mail: yxyuan@ee.cuhk.edu.hk).}
\thanks{Jun Liu is with the Department of Radiology, Second Xiangya Hospital, Changsha 410011, China. (e-mail: junliu123@csu.edu.cn).}
\thanks{Gang Li is with the Department of Radiology at the University of North Carolina at Chapel Hill, Chapel Hill 27599, USA. (e-mail: gang\_li@med.unc.edu).}
\thanks{Pingkun Yan is with the Department of Biomedical Engineering at Rensselaer Polytechnic Institute, Troy, New York 12180, USA. (e-mail: yanp2@rpi.edu).}
\thanks{Wei Liu is with the Department of Radiation Oncology, Mayo Clinic, Scottsdale 85259, USA. (e-mail: liu.wei@mayo.edu).}
\thanks{Dinggang Shen is with the School of Biomedical Engineering, ShanghaiTech University, Shanghai 201210, China; Shanghai United Imaging Intelligence Co., Ltd., Shanghai 200230, China; Shanghai Clinical Research and Trial Center, Shanghai, 201210, China. (e-mail: Dinggang.Shen@gmail.com).}}

\maketitle

\begin{abstract}
Large-scale Artificial General Intelligence (AGI) models, including Large Language Models (LLMs) such as ChatGPT/GPT-4, have achieved unprecedented success in a variety of general domain tasks. Yet, when applied directly to specialized domains like medical imaging, which require in-depth expertise, these models face notable challenges arising from the medical field's inherent complexities and unique characteristics. In this review, we delve into the potential applications of AGI models in medical imaging and healthcare, with a primary focus on LLMs, Large Vision Models, and Large Multimodal Models. We provide a thorough overview of the key features and enabling techniques of LLMs and AGI, and further examine the roadmaps guiding the evolution and implementation of AGI models in the medical sector, summarizing their present applications, potentialities, and associated challenges. In addition, we highlight potential future research directions, offering a holistic view on upcoming ventures. This comprehensive review aims to offer insights into the future implications of AGI in medical imaging, healthcare, and beyond.

\end{abstract}


\section{Introduction}
\label{sec:introduction}

In recent years, Artificial General Intelligence (AGI) models have demonstrated success in general domains. In general, AGI includes the advancement of Large Language Models (LLMs) \cite{radford2019language}, Large Vision Models \cite{kirillov2023segment}, foundation models for other modalities such as time series \cite{liang2024foundation}, and domain-specific (vertical) foundation models such as in biomedical applications \cite{zhang2024generalist} and single-cell multi-omics \cite{cui2024scgpt}. In medical image analysis, we have witnessed tremendous growth in the development and application of AGI, leading to disruptive changes in methodology design, data collection, and evaluation criteria in this field.

However, applying AGI methodologies directly to healthcare data, even with the domain-specific models, can pose significant challenges, potentially leading to decreased performance or even making it impossible \cite{cascella2023evaluating}. These challenges primarily stem from the medical domain's unique characteristics, including the specialized nature of the clinical text and medical imaging, the expertise needed for accurate interpretation, and the lack of annotation/ground truth in data. For example, medical imaging data usually encompasses images obtained from multiple modalities \cite{rajpurkar2022ai} such as X-rays, computed tomography (CT), magnetic resonance imaging (MRI), ultrasound, microscope, and much more, for the diagnosis and treatment planning of a patient's single visit  \cite{moor2023foundation}. Interpreting these images demands specialized knowledge and expertise in anatomy, pathology, and radiology, while very limited patient information of that visit (e.g., exact diagnosis, mid/long-term mortality, treatment outcome) is available. In contrast, natural images typically consist of everyday objects that can be accurately recognized by most individuals based on common sense. This distinction presents a notable challenge when it comes to annotating/chart reviewing a substantial volume of high-quality medical imaging data, as it necessitates the involvement of numerous experts, which is often impractical or unfeasible. Consequently, the medical domain faces a scarcity of extensive training data required for directly training large models from scratch, in contrast to the abundant availability of such data in the general domain. Moreover, due to the heterogeneity between medical data and general data, it is usually not optimal to directly apply AGI models trained on general data to the medical domain. In addition, clinical data usually contains sensitive patient information. Strict regulations, such as HIPAA (Health Insurance Portability and Accountability Act) \cite{Fares23} in the United States, govern the storage, transmission, and handling of patient data. As a result, training and utilizing AGI models across institutions are challenging or even impossible (commonly known as "institutional barriers") \cite{rieke2020future}. In summary, adapting AGI models to medical imaging analysis requires careful consideration due to the unique challenges of medical data.

In response, there have been various works developed to address these challenges in three major directions: data, knowledge, and models. Problems such as data scarcity and sensitivity in medical imaging have been tackled using augmentation \cite{su2023rethinking}, transfer learning \cite{subramanian2024towards}, and prompt-based methods \cite{zu2024embedded}. Gaps between expert and general domains are bridged by incorporating expert knowledge into the learning \cite{shi2023chatgraph} and expert-machine alignment processes of AGI models \cite{ge2024openagi}. Novel model design principles, including parameter-efficient AGI models \cite{zhang2024generalist} and privacy-preserved schemes \cite{kuang2024federatedscope}, are explored to better adapt to medical practice. These efforts aim to revolutionize healthcare by leveraging AGI capabilities to provide more personalized and efficient care while addressing resource scarcity in the medical field.

In this review, we begin by elucidating the technical underpinnings of AGI methodologies, detailing their primary characteristics and implementations in Section~\ref{background}. Subsequently, we explore various AGI frameworks, including Large Language Models (Section~\ref{llm}), Large Vision Models (Section~\ref{lvm}), and Large Multimodal Models (Section~\ref{lmm}) with a focus on their developments and applications in medical imaging. Each section encompasses the model developmental roadmaps, current and prospective applications, associated challenges/pitfalls, and our perspectives on potential mitigation solutions. Finally, we discuss the future direction of AGI in medical image analysis and the transformative opportunities it presents in enhancing patient care, assisting physicians, and optimizing hospital management. The main contribution of this work is three-fold: 1) It serves as a comprehensive investigation of research works related to a timely topic; 2) It provides basic explanations for a series of new methods in AGI with discussions on how they are related to medical image analysis; 3) It offers insights and visions on how AGI has and will change the field of medical image analysis. To conduct this review, we searched for keywords in each section (Large Language Model, Large Vision Model, and Large Multimodal Models) of Google Scholar up to 2024 to identify the publications that might be suitable for inclusion. We further refined the list by considering the quality and fitness of the publications. Review and perspective papers are generally not included but could be cited for their findings and opinions.

\section{Enabling Techniques of AGI Models}
\label{background}
To facilitate a comprehensive understanding of the technical content discussed in this paper, we start with an overview of the enabling technologies of AGI. The development of AGI is anchored on several technological foundations and innovations. This section is dedicated to elucidating these crucial technological underpinnings.

\begin{figure}[ht]
\begin{center}
\includegraphics[width=1\linewidth]{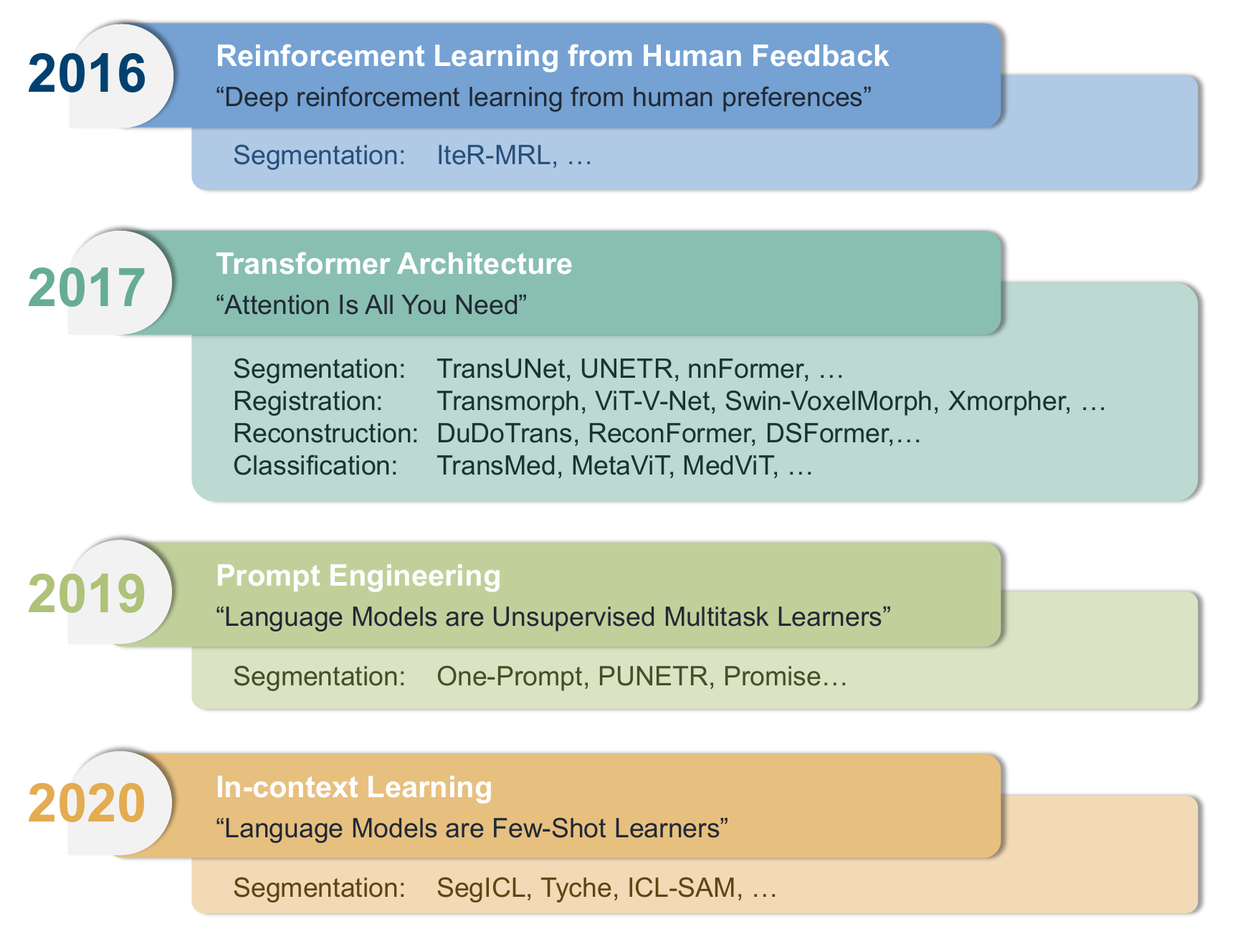}
\end{center}
\caption{Timeline of key developments in AGI techniques and their relationship to medical imaging. The timeline highlights significant milestones, starting with Reinforcement Learning from Human Feedback (2016) through the introduction of Transformer Architecture (2017), followed by advancements in Prompt Engineering (2019), and culminating in In-context Learning (2020). Each milestone is associated with various medical imaging applications such as segmentation, registration, reconstruction, and classification to showcase the evolution of AGI techniques in the medical imaging domain.}
\label{AGI_techs}
\end{figure}
\begin{itemize}[leftmargin=10pt]
  \item \textbf{Transformer Architecture}: The Transformer model, introduced in 2017 by Vaswani et al. \cite{vaswani2017attention}, has revolutionized the field of Natural Language Processing (NLP) and has become the standard neural network architecture of LLMs, LVMs, and other foundation models. At the core of the Transformer is the multi-head attention mechanism, which allows the model to assign different weights to tokens based on their relevance. This attention mechanism enables the Transformer to capture long-term dependencies more effectively. The model consists of an encoder and a decoder, each composed of multiple layers of self-attention and feed-forward neural networks. In the self-attention layer, each input token attends to all other tokens in the sequence, allowing the model to capture contextual relationships regardless of their positional distance. The multi-head aspect of the attention mechanism allows the model to attend to information from different representation subspaces at different positions, providing a more comprehensive understanding of the input. By attending to different parts of the input sequence, the Transformer can effectively understand the relationships between input tokens and capture contextual information. The model also incorporates positional encodings to maintain sequence order information, as the attention mechanism itself is position-agnostic. The combination of these elements – self-attention, multi-head mechanism, and positional encodings – enables the Transformer to process input sequences in parallel, leading to significant improvements in training efficiency and performance. The superiority of the Transformer architecture lies in its ability to address the limitations of prior models, such as Recurrent Neural Networks (RNNs) \cite{mikolov2010recurrent}, in handling variable-length sequences and context awareness. Unlike RNNs, the Transformer model does not rely on sequential processing, making it more efficient and suitable for parallel computation. This parallelism enables the Transformer to process and understand large amounts of text simultaneously, significantly improving its training and inference speed. The parallelizable nature of the Transformer architecture also makes it highly adaptable and scalable. This flexibility enables large-scale pre-training, where the model is trained on vast amounts of data to learn general language patterns and representations. These pre-trained models can then be fine-tuned for specific downstream tasks, such as text generation or language translation. This approach allows the Transformer to leverage its learned knowledge and generalize well to various domains and tasks. While deep learning models can also feature many layers of a large number of parameters, the Transformer's design allows models to take precedence over inductive biases by fully benefiting from the large size of the network to scale up the training \cite{yang2024harnessing}.
  \item \textbf{In-context Learning}: In-context learning is a powerful technique to learn and make model inferences based on a limited context provided by the user \cite{chan2022data}. Unlike traditional approaches that heavily rely on extensive pre-training or fine-tuning using large labeled datasets, in-context learning allows models to leverage specific examples or instructions within the input prompt to guide their behavior and generate contextually relevant outputs. The key idea of in-context learning is to learn from analogy. To facilitate this process, in-context learning starts by constructing a demonstration context using a few examples. These examples showcase the desired input-output behavior that the model should learn. Subsequently, the demonstration context is combined with a query question, resulting in a prompt that contains both the query and the demonstration context. This prompt is then fed into the AGI model to generate the output. In contrast with supervised learning, where model parameters are updated through backpropagation during a training stage, in-context learning does not involve explicit parameter updates. Instead, it relies on pre-trained language models that directly perform predictions based on the provided examples and prompts. The model is expected to discern the underlying patterns within the demonstration context and generate accurate predictions accordingly. With the remarkable scaling of model size and the availability of large-scale corpora, LLMs have demonstrated their proficiency in in-context learning. These models possess the capability to extract meaningful information from a few examples within the given context. By leveraging the extensive knowledge encoded in their pre-trained weights, LLMs excel at learning from limited context and making informed predictions. 
  \item \textbf{Support for Prompts}: Traditionally, the process of collecting and labeling responses for training or fine-tuning models has been both time-consuming and costly. However, prompt engineering \cite{liu2023pre} offers a more efficient alternative. A prompt, in this context, refers to a set of instructions that customize the behavior and guide the subsequent interactions and outputs of an LLM. Recent studies have demonstrated the effectiveness of prompt engineering in adapting large-scale pre-trained language models to specific downstream tasks, eliminating the need for extensive fine-tuning. Prompt engineering allows for customization and tailoring of LLM capabilities. By designing appropriate prompts, developers can shape the LLM’s behavior to align with specific objectives, tasks, or user requirements. This level of customization enhances the practical usability of LLMs across various domains and use cases. In general, prompt engineering plays a vital role in current LLMs and provides a promising way for developing advanced AGI systems. By designing appropriate prompts, developers can leverage the power of pre-trained models and optimize their performance, making LLMs and AGI systems more effective, efficient, and adaptable in the specific domain. 
  \item \textbf{Alignment to Human Feedback}: The alignment of an AGI system is a critical process that ensures its behavior is in line with desired principles, values, and objectives. This alignment is essential to ensure that AGI systems operate ethically, responsibly, and following human intentions and societal norms. While traditional supervised models can be aligned to human input via learning from labeled data, the utilization of prompts and the generalizable capability of AGIs in handling a wide variety of tasks makes them capable of more effectively learning from human feedback during inference. The most commonly adopted scheme for human-machine alignment is through Reinforcement Learning from Human Feedback (RLHF) \cite{ouyang2022training}. RLHF involves training LLMs by incorporating human feedback to guide their learning process. RLHF has been employed in LLMs such as InstructGPT \cite{ouyang2022training} and ChatGPT, to train these models by incorporating feedback from human supervisors. By leveraging human knowledge and preferences, RLHF ensures that the learned behaviors of LLMs align with human values and requirements. In traditional reinforcement learning, an agent learns through trial and error, receiving rewards or penalties based on its actions \cite{li2016developmental}. RLHF takes this process further by actively involving humans in the training loop. Human supervisors play a pivotal role by providing feedback to the learning agent, shaping its behavior, and guiding it toward desired outcomes. This approach is particularly valuable in domains where human expertise and intuition are essential, such as complex decision-making tasks or domains with limited or imperfect simulations. RLHF is not only pertinent to LLMs but also an important strategy in the development of AGI. It contributes to achieving alignment between AGI systems and human values. By incorporating human feedback, RLHF helps ensure that AGI systems learn and operate in a manner that aligns with human intentions and desires.  
\end{itemize}
In the medical imaging domain, the enabling techniques and innovations of LLMs and AGI have played a pivotal role in advancing key tasks such as segmentation, registration, reconstruction, and classification. Fig. 1 summarizes the timeline of key developments in AGI techniques and their relevance to medical imaging. Reinforcement Learning from Human Feedback (RLHF), for instance, allowed AGI systems to incorporate human preferences and expertise into the learning process, improving tasks like interactive medical segmentation (e.g., IteR-MRL~\cite{liao2020iteratively}). The Transformer Architecture has been extensively adopted across various medical imaging applications, driving significant progress in segmentation~\cite{hatamizadeh2022unetr,zhou2023nnformer}, registration~\cite{chen2022transmorph,chen2021vit,zhu2022swin,shi2022xmorpher}, reconstruction~\cite{guo2023reconformer,yu2023dsformer}, and classification~\cite{dai2021transmed,zhao2023metavit,manzari2023medvit} tasks. Prompt Engineering has further enhanced medical image segmentation by leveraging prompts to guide models toward more accurate segmentation results~\cite{wu2024one,fischer2024prompt,li2024promise}. In-context Learning has enabled models to learn from a small set of samples, eliminating the need for training the model from scratch or fine-tuning, which can be utilized for medical image segmentation tasks~\cite{hu2024icl,rakic2024tyche}.
\section{Large Language Models for Medical Imaging}
\label{llm}
Large language models are currently at the forefront of AGI development, representing the zenith of innovations in natural language processing and understanding. An earlier study by \cite{yang2024harnessing} offered a comprehensive summary of the development and categorization of large language models using an evolutionary tree. The rapid evolution of LLMs is also bringing a transformative revolution in the realm of medical imaging. Medical practitioners and healthcare organizations are now envisioning a future where AI-powered systems such as LLMs assist in disease diagnosis, patient outcome prediction, medical education, and streamlining administrative work. In this section, we'll explore the roadmaps for integrating LLMs into healthcare, discuss their present and potential roles, and highlight the challenges and obstacles of their incorporation.
\subsection{Roadmaps for LLMs in healthcare}
\begin{itemize}[leftmargin=10pt]
    \item \textbf{Expert-in-the-loop}: The incorporation of expert knowledge is vital for harnessing the full potential of LLMs within specialized medical fields. While LLMs like ChatGPT and its successors have demonstrated proficiency in general medical knowledge \cite{kung2023performance}, they might encounter challenges when performing domain-specific tasks due to a deficiency in specialized knowledge \cite{10.3389/fonc.2023.1219326,sallam2023chatgpt} and generate erroneous content. For example, in cardiology, an AI model might be tasked with interpreting electrocardiogram (ECG) readings \cite{siontis2021artificial}. While the model could identify basic patterns, accurately diagnosing complex cardiac conditions like myocardial infarctions or arrhythmias necessitates the expertise of a cardiologist. In psychiatry, a model might struggle to differentiate between overlapping mental health conditions based on patient-reported symptoms, demanding the nuanced understanding of a psychiatrist \cite{bzdok2018machine}. In radiology, distinguishing between a benign lesion and a malignant tumor on a CT scan requires the specialized knowledge that a radiologist provides \cite{van2011computer}. Likewise, in radiation oncology, determining the precise dosage and target area for radiation therapy calls for insights from radiation oncologists \cite{unkelbach2018robust}. A promising approach to overcome these obstacles is the adoption of Reinforcement Learning with Expert Feedback (RLEF). LLMs like ChatGPT are trained with RLHF \cite{ouyang2022training}, where generated responses during training are ranked to enhance and refine the model's outputs. Similarly, by integrating radiologists' specialized knowledge within the feedback loop, AGI models can mirror the nuances of diverse medical specialties more accurately, leading to clinically relevant and precise outputs.

\begin{figure}[t]
\label{figure4}
\begin{center}
\includegraphics[width=0.45\textwidth]{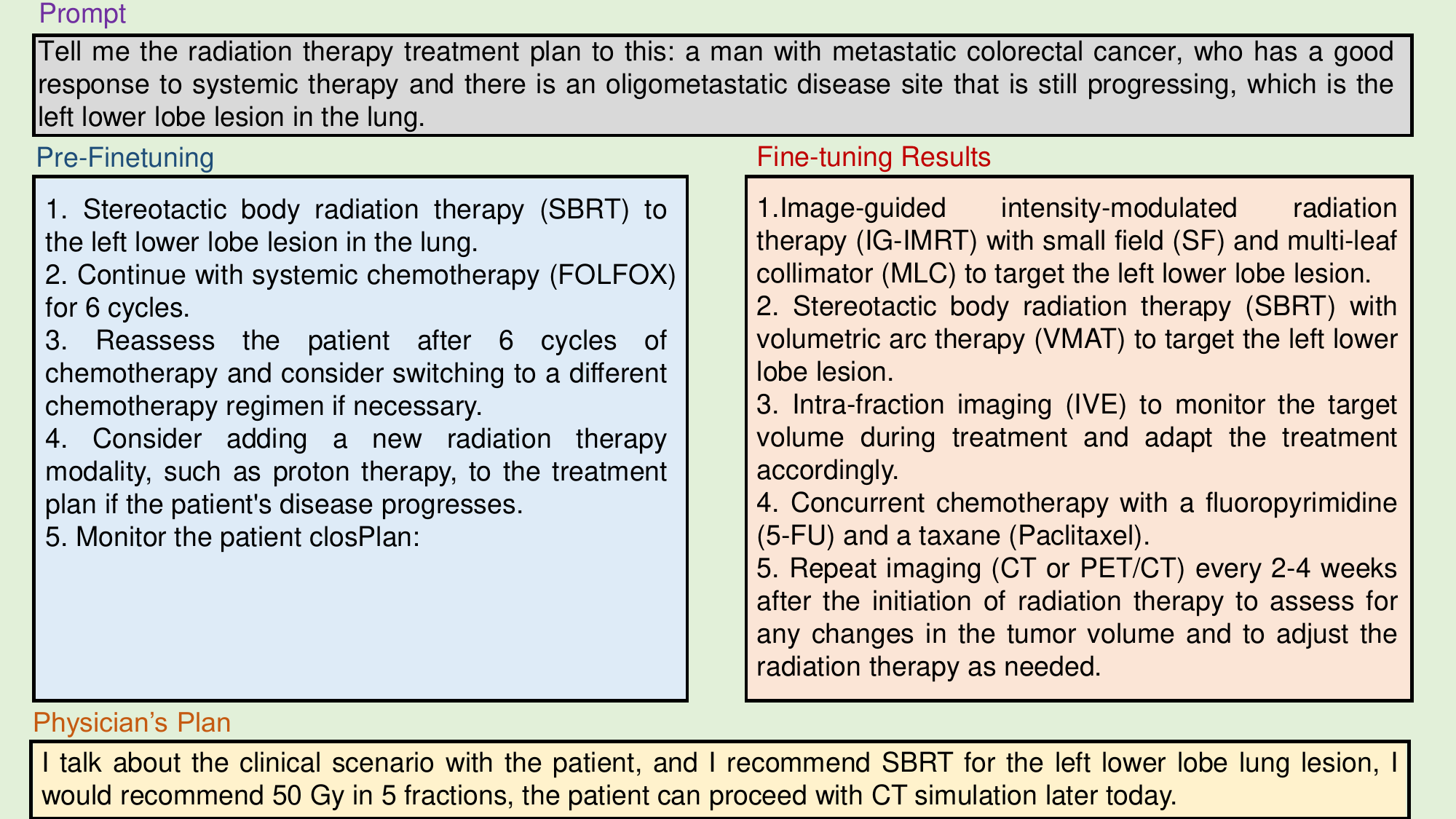}
\end{center}
\caption{RadOnc-GPT: world's first domain-specific LLM fine-tuned for radiation oncology. It digests patient information and generates radiation therapy treatment regimens \cite{liu2023radonc}.}
\end{figure}

    \item \textbf{Medical Domain Adaptation}: Adapting language models for specialized fields, especially medicine, requires careful consideration and domain-specific expertise \cite{gu2021domain}.
    A pivotal facet of domain adaptation for LLMs is harnessing the medical domain-specific data. Medical data, although scarce, is a concentrated source of insights. When incorporated, it equips LLMs to navigate the nuanced labyrinth of medical terminology and diagnostic subtleties. By honing the models with this targeted information, they're transformed into powerful tools that can offer precise, actionable insights, enhancing the efficiency and accuracy of healthcare professionals in their practice. Prompt tuning is an effective and low-cost approach to adapt Large Language Models (LLMs) to specific domains, data, and tasks \cite{liu2023pre} without updating the massive model parameters of large foundational LLMs. In the context of the medical domain, this can be particularly powerful. For instance, when dealing with intricate medical terminologies or complex diagnostic scenarios, a well-tuned prompt can act as a guidepost, directing the LLM to retrieve or generate relevant and precise information. This can be invaluable in situations like differential diagnoses, where slight nuances in language can lead to vastly different clinical interpretations. Additionally, prompt tuning can serve as a bridge to merge the vast knowledge stored within LLMs with the specialized requirements of the medical field. Instead of undergoing resource-intensive retraining, the LLM can be steered toward medical accuracy through carefully calibrated prompts. This not only saves computational resources but also time, making the LLM readily adaptable for urgent medical applications.

    

    \item \textbf{Integrating Medical Informatics}: The successful integration of LLMs, such as ChatGPT, into clinical workflows is fundamentally dependent on the robustness of medical informatics. The swift, efficient, and precise collection and integration of data is integral for effectively deploying or updating these models in healthcare environments. As a result, there is a burgeoning need to investigate approaches for sourcing textual or other pertinent data from complex \cite{aggelidis2008methods} hospital information systems. This access to data is imperative for incorporating LLMs or multi-modal models into areas such as clinical diagnostics, treatment, radiation oncology, and radiology. In this context, informatics plays a vital role, serving as a crucial bridge connecting vast, intricate data sets with AGI applications in healthcare. The discipline's function in enabling data accessibility, deciphering it, and applying it subsequently is invaluable in the forward march of AI in clinical environments. This emphasizes the call for scientific exploration to pivot towards the further refinement and enhancement of medical informatics, thereby nurturing a more potent relationship between this field and AGI. This harmonious integration promises not only to boost the operational efficiency of healthcare systems but also to augment patient outcomes via more precise and prompt diagnoses and treatments.

\begin{figure*}[!ht]
\begin{center}
\includegraphics[width=0.9\textwidth]{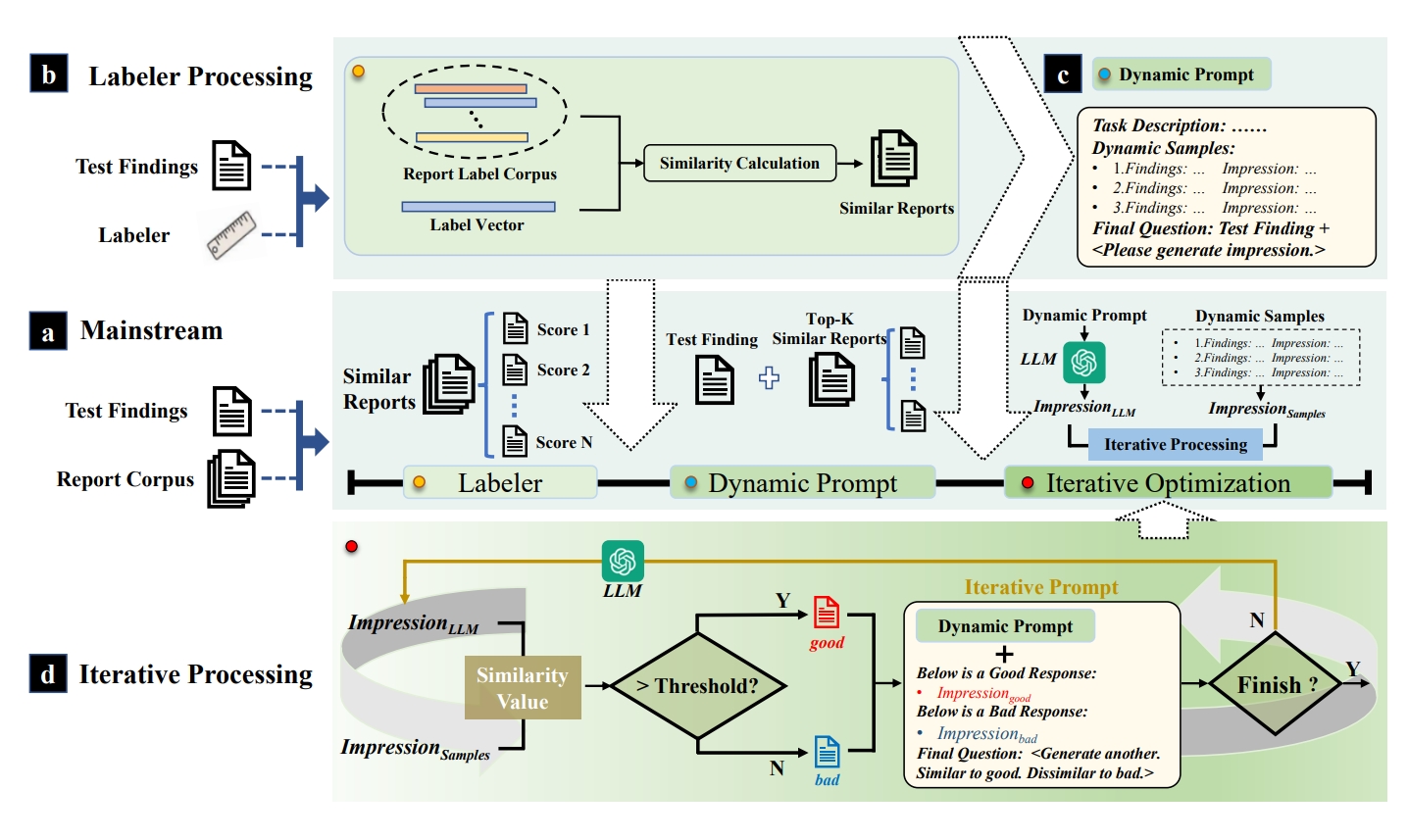}
\end{center}
\caption{The pipeline of ImpressionGPT. Part (a) in the middle panel is the mainstream of the ImpressionGPT method. The authors use a labeler to categorize the diseases of test report and obtain similar reports in the corpus (part b), and then construct a dynamic prompt in part c. Part d accomplishes the iterative optimization of LLM through interaction with positive (good) and negative (bad) responses}
\label{impression}
\end{figure*}
    
    \item \textbf{Knowledge Graph}: In the nexus of healthcare and technology, knowledge graphs stand as powerful allies to LLMs like ChatGPT. These graphs, acting as structured repositories of interconnected medical data, can enrich the input and prompts to LLMs \cite{brate2022improving} and provide a means to verify its outputs \cite{peng2023check}. Instead of merely generating text based on patterns, when paired with a medical knowledge graph, an LLM can reference a vast, organized network of medical insights. Such a setup ensures that the AI's responses are not just fluent but also contextually accurate and medically sound. Furthermore, with the automation of knowledge extraction processes, the creation and refinement of knowledge graphs can be expedited. This would consequently enhance the accuracy, speed, and efficacy of LLMs in healthcare.    
    \item \textbf{Collaboration}: Collaborations are pivotal in the deployment and optimization of LLMs in healthcare, particularly in addressing complex issues such as data diversity and data privacy. Effective modeling of medical data requires careful consideration of patient demographics, socio-economic diversity, and cultural factors \cite{jacoba2023bias}. It is necessary to develop models that are trained and adapted to data representative of diverse hospitals and institutions across a nation or region.
    
\end{itemize}

\subsection{Current and Potential Applications}


\begin{itemize}[leftmargin=10pt]
\item \textbf{Medical Report Understanding and Generation}:
LLMs, like ChatGPT, hold significant potential in medical report understanding and generation by effectively reasoning with structured input data such as symptoms, test results, and medical history to produce coherent and relevant reports. Wu et al. conducted a comparative study examining general LLMs such as ChatGPT against domain-specific fine-tuned models in performing natural language inference tasks within the radiology domain \cite{wu2023exploring}. Results reveal that LLMs exhibit a robust ability to understand and reason medical reports while maintaining zero-shot or few-shot data efficiency. This is particularly notable as they achieve these results without fine-tuning on labeled data, which is often limited in the medical domain. An illustrative report generation example is ImpressionGPT which uses the ChatGPT to generate the Impression section based on the Findings section of a radiology report \cite{10433180}, which can be laborious and error-prone for radiologists. As illustrated in Fig. 3, ImpressionGPT leverages the in-context learning capability of LLMs by constructing dynamic contexts using domain-specific, individualized data. This dynamic prompt approach enables the model to learn contextual knowledge from semantically similar examples from existing data. ImpressionGPT achieves state-of-the-art performance on both MIMIC-CXR and OpenI datasets without requiring additional training data or fine-tuning the LLMs\cite{10433180}.

\item \textbf{Disease Diagnosis and Patient Outcome Prediction}: 
LLMs such as ChatGPT offer promising capabilities in determining the cause of diseases and predicting patient outcomes. By harnessing the power of LLMs, we can transform text data into a suitable format for predicting outcomes and diagnosing patients exhibiting similar symptoms. This concept can be exemplified by feeding case studies of patients with identical symptoms to ChatGPT to predict potential disease outcomes based on patterns from the provided data.


\item \textbf{Application in Medical Education and Patient Consultation}:
LLMs could be an instrumental tool in educating the next generation of healthcare practitioners and even inpatient consultations \cite{lee2023rise}. For instance, medical students could interact with ChatGPT for scenario-based learning, and patients may use it as an accessible source of information to better understand their medical conditions. Another scenario for LLMs is medical dialogue summarization, such as the dialogue during the patient consultation.

\item \textbf{Streamlining Clinical and Administrative Work}:
LLMs such as ChatGPT and GPT-4 could help physicians with writing clinical notes \cite{liu2023summary} or producing radiology reports \cite{10433180}. In addition, LLMs can significantly reduce the burden of clinical administrative tasks such as drafting administrative documents. freeing up physicians' time for more important tasks in patient care.

\item \textbf{Unified Knowledge Space}:
Expressing information in a unified space that encompasses text, images, and genetic data can revolutionize fields such as bioinformatics and genomics. By creating a unified knowledge graph, relevant information from diverse sources can be mapped into a graphical space, which could potentially change the field of bioinformatics, clinical informatics, medical imaging, and genomics. A consolidated sphere of knowledge can serve as a springboard for breakthroughs. For example, future models can leverage this vast data repository. They could comprehensively analyze a patient's textual clinical history, diagnostic imaging findings, laboratory test outcomes, and individual genetic information. In doing so, they would be equipped to render more insightful, well-rounded decisions, thereby improving patient outcomes.

\item \textbf{Integrating text data and other modalities}:
The significance of integrating different types of data into a unified space for efficient decoding and analysis is pivotal across various medical domains. In cardiology, for example, text-based patient reports, imaging data from ultrasound or MRIs, audio data like heart sounds, and echocardiogram (ECG) signals could be organized as multimodal inputs to LLMs. An advanced language model could then analyze these multimodal inputs to provide a more comprehensive and accurate diagnosis of heart conditions, thus supporting informed clinical decision-making. Similarly, in telemedicine and mental health, the amalgamation of patient-reported textual data, audio-visual inputs, wearable device metrics, biometric data, and even environmental sensor data is beneficial. LLMs such as Kosmos-1 \cite{huang2023language}, with their capacity to process and make sense of such multimodal inputs, can potentially revolutionize these fields by providing more accurate, personalized, and nuanced health assessments and treatment suggestions.


\end{itemize}

\subsection{Challenges and Pitfalls}
\begin{itemize}[leftmargin=10pt]

\item \textbf{Prompting Challenges}: The construction of standardized and appropriate prompt words stands as a significant hurdle for the optimal performance of LLMs like ChatGPT \cite{electronics13152961}. The effectiveness of these prompts directly influences the quality of responses, thus crafting the right prompt is as pivotal as refining the model itself.

\item \textbf{Data Privacy and Regulations}: Medical data utilization in AGI models brings about unique challenges in terms of data privacy and compliance with regulatory bodies. Protecting personal health information and meeting the stringent requirements of institution review boards (IRBs) and HIPAA guidelines \cite{murray2011privacy} is paramount in any healthcare-related AI deployment. A potential solution could be the application of local LLMs within hospitals, which may mitigate privacy concerns while allowing models to learn from a vast array of clinical data. 

\item \textbf{Data Accessibility}: Accessing medical data for AGI models raises not only technical but also legal and ethical complexities \cite{char2020identifying}. Balancing the necessity of access for model training and operation against the privacy rights of patients and healthcare providers is a complex undertaking.

\item \textbf{Preparation and Curation of Healthcare Data}: The need for image and corresponding text data in healthcare poses a significant challenge. The availability of comprehensive datasets akin to the MIMIC series \cite{johnson2019mimic} is scarce but critical for the effective training and utilization of LLMs in healthcare applications.

\item \textbf{Deployment in Local Hospitals}: The execution of LLMs locally in hospitals necessitates substantial computational resources. This could prove to be a limiting factor, particularly for smaller hospitals with limited access to high-performance computational infrastructure.

\item \textbf{Data Imbalance}: Real-world clinical data often exhibits imbalance, with disproportionate representation of certain conditions, demographics, or other variables \cite{liu2022survey}. This poses a challenge for certain models that rely on balanced, representative data for robust and unbiased outputs. 

\end{itemize}

\section{Large Vision Models for Medical Imaging}
\label{lvm}
In recent years, deep learning has made significant strides in the field of medical imaging, revolutionizing the way medical images are analyzed and interpreted. Various deep learning models, such as Convolutional neural networks (CNNs) \cite{ronneberger2015u} and Vision Transformer (ViT) \cite{dosovitskiy2020image}, have shown remarkable success in a wide range of tasks like medical image reconstruction, segmentation, and classification. Many of these models have been deployed to assist radiologists and clinicians in tasks such as identifying abnormalities, localizing tumors, and quantifying disease progression \cite{anwar2018medical}. Recently, the emergence of large vision models, such as the DINOv2~\cite{oquab2023dinov2} and Segment Anything (SAM) model \cite{kirillov2023segment}, may further propel the advancement in the medical imaging field, ultimately leading to improved patient outcomes and more efficient healthcare practices. In this section, we will delve into the roadmaps for deploying and adapting large vision models in the medical domain, explore the current and potential applications, and discuss the challenges and pitfalls associated with their integration and utilization.

\subsection{Roadmaps}

\begin{itemize}[leftmargin=10pt]
    \item \textbf{Large-Scale Datasets}: Large-scale, high-quality, and diverse medical imaging datasets play a crucial role in deploying and adapting large vision models in the medical domain. Gathering and curating such datasets require collaborations between researchers, medical professionals, and institutions. These datasets need to cover a wide range of pathologies, modalities, and patient populations to capture the full spectrum of medical conditions and ensure the generalizability of the models. Additionally, data privacy and security measures must be in place to protect patient information and comply with ethical standards. Federated learning \cite{rieke2020future,kairouz2021advances} is a promising approach for addressing privacy and data security concerns in the deployment of large vision models in the medical imaging field. By leveraging federated learning, large vision models can be trained using a vast amount of distributed medical imaging data while preserving data privacy. This approach is particularly beneficial in scenarios where data cannot be easily shared or centralized due to legal, ethical, or logistical constraints. Federated learning also provides an avenue for harmonizing diverse datasets from different institutions, capturing the variability and heterogeneity in medical imaging. This can improve the generalization capability of large vision models and their performance across multiple healthcare settings, enhancing the model's robustness and adaptability. Approaches for ensuring fair and secure use of data would also be vital for a sustainable and reliable ecosystem in medical image data usage \cite{kaissis2020secure}. 

    \item \textbf{Model Adaption}: The adaptation of large vision models to the medical domain involves fine-tuning or transfer learning techniques \cite{esteva2019guide}. Pretrained models on large-scale general image datasets, such as ImageNet \cite{deng2009imagenet}, may serve as a valuable starting point. However, to leverage their learned representations effectively, these models need to be further trained on medical imaging data. Fine-tuning requires careful optimization and regularization strategies to adapt the models to specific medical imaging tasks while preventing overfitting and preserving their generalization capabilities. Recently, adding adapters \cite{chen2023sam} to large vision models has indeed gained popularity as a flexible and efficient approach to model customization and transfer learning. Adapters allow for the integration of task-specific information without modifying the entire model architecture, enabling faster and more cost-effective model development.

    \item \textbf{Multimodal Imaging}: Medical imaging often involves multiple modalities, such as ultrasound, MRI, CT, and PET, etc., each providing unique information. The adaption of large vision models should consider the combination and fusion of information from these multiple modalities to extract complementary features and enhance the analysis. Exploring fusion techniques, such as late fusion \cite{gadzicki2020early}, early fusion \cite{gadzicki2020early}, or cross-modal attention mechanisms \cite{wang2022cross}, may be crucial to leverage multi-modality data effectively and improve the model performance.

    \item \textbf{Interpretability}: Large models often operate as complex black boxes, making it challenging to understand the reasoning behind their predictions \cite{bohle2023holistically}. To establish trust and facilitate clinical decision-making, efforts should be made to explain the decision-making process of these models. Interpretability methods such as attention maps \cite{hassanin2022visual}, saliency maps \cite{simonyan2013deep}, and Grad-CAM \cite{selvaraju2017grad} have been explored to provide insights into which regions of an image contribute most to the model's decision, aiding in understanding and verifying the model's outputs. However, the interpretability of large vision models has not been fully explored, which requires further efforts from the community.

    \item \textbf{Few-shot/Zero-shot Learning}: Few-shot and zero-shot learning hold significant potential in the medical imaging field, where obtaining extensive annotated datasets can be time-consuming, expensive, and even infeasible \cite{tajbakhsh2021guest}. LLMs, such as GPT-3 \cite{brown2020language} and its successors \cite{liu2023summary}, have demonstrated impressive zero-shot learning capabilities, where they can generate meaningful responses or perform tasks without explicit training on specific examples. Translating this concept to the medical imaging domain, large vision models can potentially \cite{kirillov2023segment} exhibit zero-shot learning abilities, allowing them to recognize and analyze new medical conditions or imaging modalities for which no labeled training data is available. This capability could enable them to adapt to previously unseen diseases, imaging techniques, or even cross-modal tasks. For example, a large vision model trained on a diverse range of medical imaging data could potentially infer and interpret new types of images or identify rare conditions, based on the analogies and knowledge learned from similar cases.

    \item \textbf{Scalability}: Training and deploying large vision models demand significant computational resources, including high-performance computing infrastructure and efficient parallel processing capabilities. Real-time applications may require optimizations to meet the time constraints of clinical settings. For example, real-time image analysis is required for rapid decision-making in emergency situations or during surgical procedures. The scalability of large vision models is crucial to ensure their practical applicability and efficiency in these time-sensitive scenarios. To address the scalability challenge, advancements in hardware acceleration, such as specialized graphical processing units (GPUs) \cite{hagerty2017medical} or tensor processing units (TPUs) \cite{jouppi2018motivation}, can significantly boost the computational efficiency of large vision models. Parallel processing techniques \cite{garland2008parallel}, distributed computing \cite{kaissis2020secure}, model compression \cite{choudhary2020comprehensive}, and distillation methods \cite{polino2018model} can also enhance scalability by optimizing memory utilization and reducing computational overhead.
    
\end{itemize}

\subsection{Current and Potential Applications}

\begin{figure}[t]
\begin{center}
\includegraphics[width=0.45\textwidth]{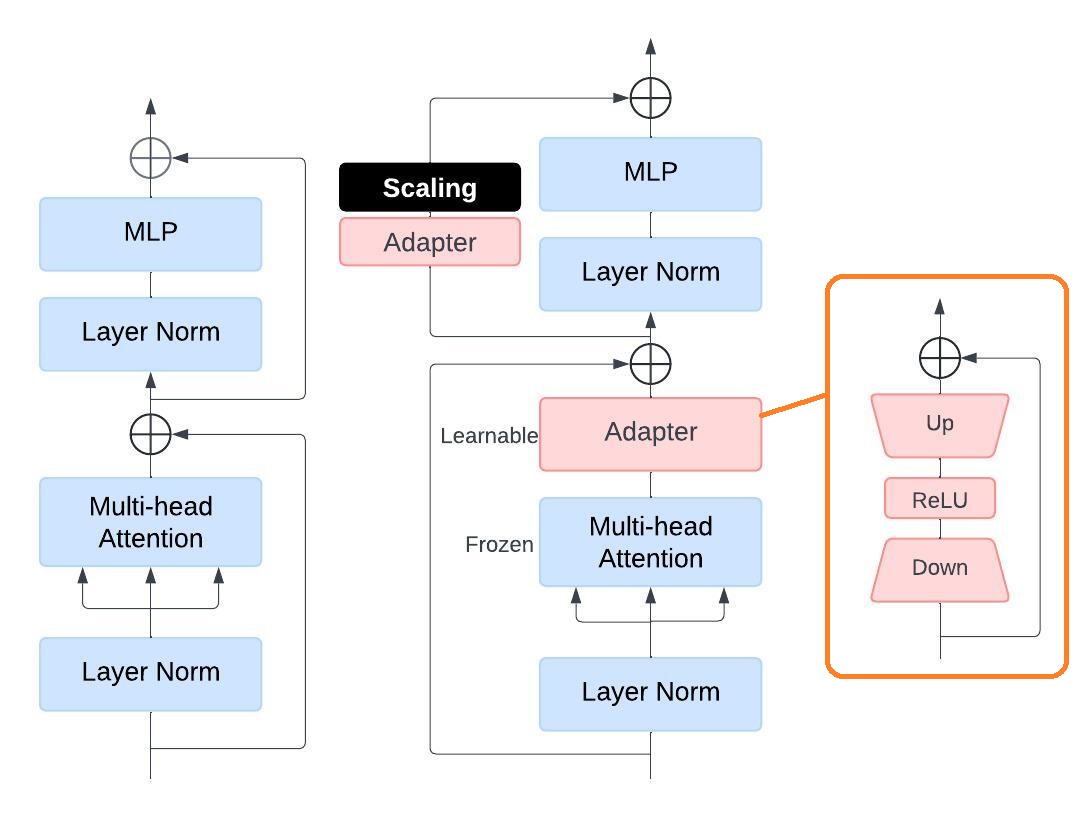}
\end{center}
\caption{ViT block in SAM and the 2D medical image adaption of SAM proposed in \cite{wu2023medical}.}
\end{figure}

\begin{itemize}[leftmargin=10pt]
    \item \textbf{Segment Anything}: Large vision models have demonstrated significant potential in various medical imaging applications, enabling improved diagnostic accuracy, treatment planning, and disease monitoring. One notable large vision model is the Segment Anything (SAM) model \cite{kirillov2023segment}, which was trained on the SA-1B dataset with over 1 billion masks on 11 million images. SAM supports promptable segmentation over various segmentation tasks and demonstrates impressive and powerful zero-shot generalization ability. Recently, SAM has attracted a lot of attention in the medical imaging field, and a group of studies explored SAM's ability in medical image segmentation tasks. For example, He et al. evaluated the SAM's accuracy on 12 medical image segmentation datasets, showing that SAM underperforms the state-of-the-art methods in these datasets and suggesting the necessity to adapt SAM for medical imaging application \cite{he2023accuracy}. To improve the performance of SAM on medical image segmentation, another group of studies explored adapting SAM for specific tasks \cite{ma2023segment,chen2023sam,wu2023medical,qiu2023learnable,zhang2023customized}. Ma and Wang et al. utilized a straightforward way of fine-tuning SAM to general medical image segmentation\cite{ma2023segment}. And the results indicate a great improvement compared with the default SAM model. Wu et al. proposed a Med SAM Adapter (Fig. 4) by adding an adapter to both the image encoder and mask decoder with comparable and even superior performances over state-of-the-art methods \cite{wu2023medical}. Zhang and Liu et al. customized SAM for medical image segmentation by applying the low-rank-based (LoRA) finetuning strategy to SAM \cite{zhang2023customized}. SAM has also been employed in medical image annotation. Liu et al. integrated SAM with 3D Slicer, an open-source software for visualization, segmentation, and analysis of medical images, to support interactive annotation \cite{liu2023samm}. Despite the aforementioned signs of progress, SAM has great potential in enabling precise quantitative measurements, volumetric analyses, and 3D reconstructions and supporting clinicians in making informed decisions and providing personalized patient care.
        \item \textbf{AI-Generated Content (AIGC)}: Diffusion-based Modeling and Generative Adversarial Networks (GAN) 
    AIGC aims to generate digital content based on human input (e.g., instructions and/or exemplars) \cite{cao2023comprehensive}. \cite{rombach2022high} has gained recognition for its high-performance capabilities in generating detailed images by leveraging a latent diffusion model conditioned on text. This model has exhibited remarkable efficacy in various tasks, including image inpainting and super-resolution. Additionally, the DALL·E 2 model introduces a groundbreaking ability to generate synthetic images from textual descriptions \cite{ramesh2022hierarchical}. In the context of the medical domain, AIGC models hold immense potential for addressing prevalent challenges. For example, medical images often suffer from issues such as noise, artifacts, and low contrast, which can significantly impact diagnostic accuracy. Generative modeling can be harnessed to mitigate these problems by effectively removing noise and artifacts while enhancing the overall image quality in medicla imaging scans \cite{waqas2023revolutionizing}. This enhancement facilitates clearer visualization of anatomical structures and pathologies, aiding healthcare professionals in accurate diagnosis and treatment planning. Generative modeling can be also utilized to generate synthetic medical images for training purposes \cite{lang2024using}, allowing medical professionals to simulate rare or challenging clinical scenarios and improve their diagnostic skills. Moreover, these synthetic images can be employed to augment limited datasets, providing a valuable resource for training large vision models in medical imaging.
    
\end{itemize}

\begin{table*}[h!]
\caption{Comparison of strategies for adapting SAM models to medical images}
\label{table:sam_medical_images}
\renewcommand\arraystretch{1.25}
\centering
\begin{tabularx}{\textwidth}{>{\centering}m{0.15\textwidth}|>{\centering}m{0.15\textwidth}|>{\centering\arraybackslash}m{0.55\textwidth}}
\hline
\hline
\textbf{Strategies used to adapt to medical images} & \textbf{Studies} & \textbf{Details (Methods + Results)} \\ \hline
\multirow{5}{2cm}{\centering \textbf{zero-shot}} 
    & Sheng He, et al., 2023 \cite{he2023computer} & \textbf{Results:} Accuracy in 12 datasets is much lower than U-Net-based models \\ \cline{2-3}
    & Saikat Roy, et al., 2023 \cite{roy2023sam} & \textbf{Results:} Performance in a multi-organ dataset of the CT domain is worse than U-Net-based models \\ \cline{2-3}
    & Ruining Deng, et al., 2023 \cite{deng2023segment} & \textbf{Results:} Performance in three tasks is worse than the SOTA methods \\ \cline{2-3}
    & Maciej A. Mazurowski, et al., 2023 \cite{mazurowski2023segment} & \textbf{Results:} Performance varies significantly across different datasets and image modalities in 28 tasks \\ \cline{2-3}
    & Peilun Shi, et al., 2023 \cite{shi2023generalist} & \textbf{Results:} Inconsistent zero-shot performance in nine medical image segmentation benchmarks \\ \hline
\multirow{3}{2cm}{\centering \textbf{high-quality downstream data + proper fine-tuning strategies}} 
    & Jun Ma, et al., 2023 \cite{ma2024segment} & \textbf{Downstream Dataset:} 1,570,263 image-mask pairs, covering 10 imaging modalities and over 30 cancer types. \textbf{Results:} Achieving better accuracy and robustness than modality-wise specialist models on 86 internal validation tasks and 60 external validation tasks \\ \cline{2-3}
    & Jay N. Paranjape, et al., 2023 \cite{paranjape2024adaptivesam} & \textbf{Fine-tuning:} A novel text-prompted segmentation method + a novel bias-tuning strategy. \textbf{Results:} Achieving SOTA results on three publicly available surgical scene segmentation datasets \\ \cline{2-3}
    & Can Cui, et al., 2023~\cite{paranjape2024adaptivesam} & \textbf{Fine-tuning:} Utilization of weak and few annotations. \textbf{Results:} Surpasses the SOTA methods in a nuclei segmentation task on the public Monuseg dataset \\ \hline
\multirow{3}{2cm}{\centering \textbf{specially designed architectures}} 
    & Junde Wu, et al., 2023~\cite{wu2023medical} & \textbf{Med-SA:} A novel Medical SAM Adapter (Med-SA) to incorporate domain-specific medical knowledge into the segmentation model using a light yet effective adaptation technique. \textbf{Results:} Med-SA outperforms SOTA medical image segmentation methods, while updating only 2\% of the parameters \\ \cline{2-3}
    & Shizhan Gong, et al., 2023~\cite{gong20233dsam} & \textbf{3DSAM-Adaptor:} A new parameter-efficient adaptation method to holistically adapt SAM from 2D to 3D for medical image segmentation. \textbf{Results:} Outperforms domain SOTA image segmentation models on 3 out of 4 tasks by 8.25\%, 29.87\%, and 10.11\% improvement for kidney tumor, pancreas tumor, colon cancer segmentation, and achieves similar performance for liver tumor segmentation \\ \cline{2-3}
    & Cheng Chen, et al., 2023~\cite{chen2024ma} & \textbf{MA-SAM:} A novel modality-agnostic SAM adaptation framework that is applicable to various volumetric and video medical data. \textbf{Results:} Outperforms various SOTA methods on five medical image segmentation tasks, by using 11 public datasets across CT, MRI, and surgical video data \\ \hline
\multirow{2}{2cm}{\centering \textbf{effective prompts}} 
    & Yuhao Huang, et al., 2023~\cite{huang2024segment} & \textbf{Downstream Dataset:} Collecting and sorting 53 open-source datasets and building a large medical segmentation dataset with 18 modalities, 84 objects, 125 object-modality paired targets, 1050K 2D images, and 6033K masks. \textbf{Prompts:} Incorporating various manual prompts to aid in accurate segmentation. \textbf{Conclusion:} Different prompt strategies have significant influence on the performance of the model \\ \cline{2-3}
    & Dongjie Cheng, et al., 2023 \cite{cheng2023sam} & \textbf{Downstream Dataset:} Collecting 12 different publicly available datasets covering CT, X-ray, MRI, Endoscopy, Ultrasound, and OCT. \textbf{Prompts:} Evaluating auto-prompt mode, box-prompt mode, and point-prompt mode. \textbf{Conclusion:} Even with proper prompts, the overall performance is still not as good as that of the trained model with the supervised method \\ \hline
\hline
\end{tabularx}
\end{table*}

\subsection{Challenges and Pitfalls}

A common strategy for applying large vision models to the medical imaging domain is to leverage models pre-trained on natural images and then develop advanced algorithms to tackle downstream medical tasks. However, due to the inherent differences between natural and medical images, the default performance of these models is often insufficient for direct application to medical imaging. To achieve state-of-the-art results in these specialized areas, several approaches have been proposed to enhance performance. These include fine-tuning with high-quality, domain-specific data, integrating adapters with task-specific architectures, and using carefully crafted prompts alongside manual annotations. We have summarized key studies in Table \ref{table:sam_medical_images}. While the adaptation and utilization of large vision models in the medical domain demonstrate great promise, some several challenges and pitfalls may be encountered and need to be effectively addressed. 

\begin{itemize}[leftmargin=10pt]

\item \textbf{The Availability of High-quality Annotations}: Training large vision models requires large-scale, accurately annotated datasets. However, medical imaging datasets are often limited in size and quality due to factors like data scarcity, privacy concerns, and variations in imaging modalities. The scarcity of labeled data can hinder the performance and generalizability of large vision models. To overcome this challenge, approaches such as active learning \cite{monarch2021human}, data augmentation and transfer learning from related tasks can be employed. Collaboration among researchers, healthcare institutions, and regulatory bodies is also essential to foster the sharing and creation of standardized, annotated datasets that cover a wide range of medical conditions and patient demographics.

\item \textbf{High Accuracy Standards}: One of the critical challenges in deploying large vision models in the medical imaging field is the requirement for higher accuracy standards compared to general image applications. While a certain level of accuracy may be acceptable in general computer vision tasks, there is a higher demand for accuracy, reliability, and precision in medical imaging applications where decisions regarding diagnosis and treatment are made based on the results provided by these models. A lower classification accuracy or an inaccurate segmentation in medical images could lead to misdiagnosis, incorrect treatment plans, delayed interventions, or missed critical findings, potentially compromising patient safety and well-being.

\item \textbf{Long-tail Problem}: Another significant challenge that large vision models face is the long-tail distribution of diseases and conditions. The long-tail phenomenon refers to the uneven distribution of labeled data, where a few common diseases or conditions have abundant training examples, while a vast majority of rare or less prevalent conditions have limited labeled data available. This long-tail challenge poses difficulties for large vision models as they may struggle to accurately classify and diagnose rare or uncommon medical conditions due to limited exposure during training. The lack of sufficient labeled examples for these conditions can result in a performance bias towards the more prevalent diseases, leading to suboptimal accuracy and sensitivity in real-world clinical settings.

\item \textbf{Ethical Issues}: The ethical implications of using large vision models in healthcare must be carefully considered. Issues such as data privacy, patient consent, and potential biases in the models require attention. Large vision models heavily rely on extensive datasets for training, which raises concerns about the privacy and security of patient information, as well as vulnerabilities of the healthcare systems to cybersecurity hazards such as backdoor attacks \cite{299844}. Strict data governance policies, anonymization techniques \cite{10630540}, and adherence to regulatory frameworks such as HIPAA are crucial to protect patient privacy. Furthermore, biases present in the training data \cite{cirillo2020sex,liu2023summary} can inadvertently propagate into the predictions of large vision models, leading to disparities in healthcare outcomes. Addressing and mitigating these biases is vital to ensure fair and equitable deployment across diverse patient populations.



\end{itemize}

In addition to the challenges mentioned above, there are also other notable challenges in deploying large vision models in the medical imaging field, such as the interpretability of the large vision models, the deployment in clinical scenarios, etc. In conclusion, while leveraging large vision models for medical imaging holds tremendous potential, addressing the aforementioned challenges is essential for realizing the full benefits of large vision models and ensuring their safe, effective, and ethical integration into clinical practice, and leading to improved healthcare outcomes and enhanced patient care.

\section{Large Multimodal Models for Medical Imaging}
\label{lmm}

Data in the real world are captured across various modalities such as text, images, videos, and audio, mirroring the perceptual capabilities of the human brain. In the medical field, data encompassing medical reports, clinical notes, radiology images, physician dictations, audio recordings of physician-patient dialogues, and surgical videos \cite{johnson2016mimic,johnson2019mimic,zeng-etal-2020-meddialog,ben2019question} paint a comprehensive picture of the operational processes of the hospital. Typically, these data can be consolidated at the patient level as they all pertain to specific patients and/or diseases, paving the way to leverage these rich data for a broad range of applications in the medical domain. Recently, rapid development has been observed in large multimodal models, fueled by the emergence of LLMs and expansive vision models. Newly released models, such as Meta's Segment Anything model \cite{kirillov2023segment}, OpenAI's DALL·E 2 \cite{ramesh2022hierarchical}, and Stability AI's Stable Diffusion \cite{rombach2022high} model, all exhibit text-to-image capabilities to prompt segmentation or generation tasks. As a result, we will mainly focus on Vision Language models in the medical domain in this section. Specifically, we will describe the roadmap for deploying and adapting large multimodal models in the medical domain, introduce current and potential applications, and discuss challenges that could further catalyze the use of large multimodal models in the medical domain.

\subsection{Roadmaps for large multimodal model}

\begin{itemize}[leftmargin=10pt]
    \item \textbf{Scaling and Quality of Paired Data}: In the medical field, paired multimodal data are essential for adapting large multimodal models; however, they are never sufficient for data-hungry large-scale models, regardless of whether they are in the general or medical domain. In the general domain, it is relatively easy to collect text-image pairs from web images that are accompanied by contextual content or human-annotated captions, which don't necessitate specialist knowledge for captioning \cite{radford2021learning}. The newly released multimodal dataset in a general domain such as LAION-5B \cite{NEURIPS2022_a1859deb} contains billions of text-image pairs. Conversely, in the medical domain, multimodal datasets are significantly smaller compared to the general domain, reducing from billions to thousands, exemplified by MIMIC-CXR \cite{johnson2019mimic} and Open-I \cite{demner2016preparing}, with 227,835 and 3,955 radiology image-report pairs respectively.
    
    Despite their considerably smaller scale in comparison to the general domain, the quality of multimodal data in the medical domain is inherently superior. For instance, in MIMIC-CXR, radiology reports are accurately and concisely written by radiologists to describe the findings and impressions from images. As a tradeoff, the amount of multimodal data in the medical domain is very limited, as it requires experts with medical knowledge to annotate and must adhere to certain privacy rules before being released for public use.

    \item \textbf{Pretraining Multimodal Foundational Models}: Given the fact that multimodal data in the general domain consists of a relatively large amount of data but with sparse information and varying quality, while data in the medical domain is limited but of high quality, an efficient and effective approach to deploying and adapting a large multimodal model in the medical domain is to first pre-train a foundation model on large-scale general domain dataset and then fine-tune the pre-trained foundation model on high-quality medical dataset. Leveraging the zero-shot in-context learning abilities of pre-trained foundational models, such as OpenClip \cite{ilharco2021openclip} for image and text retrieval and Segment Anything Model for image segmentation, we can feasibly adapt these models to the medical field. They can be readily used for tasks such as medical report generation, radiology image comprehension, and medical image segmentation, even when only limited fine-grained medical multimodal data is available.

    \item \textbf{Parameter-Efficient Fine-Tuning}: Recently released foundation models possess two primary characteristics: 1) a large scale of parameters and 2) large pre-training datasets. For instance, GPT-3 \cite{brown2020language} has 175 billion parameters trained on 45 TB of data. LLaMA \cite{touvron2023llama} models range from a scale of 7 B to 65 B trained on 1.4 trillion tokens. This expansive scale leads to high computational costs for full fine-tuning. Hence, Parameter-Efficient Fine Tuning (PEFT) strategies \cite{dutt2023parameter,chen2023parameter} can serve as a potent method to rapidly adapt these pre-trained foundation models to the medical domain. Below are some prevalent PEFT strategies for tuning large pre-trained models: Adapter tuning \cite{houlsby2019parameter} introduces lightweight modules, referred to as "adapters", within the layers of the pre-trained large models. When these models are fine-tuned on downstream tasks, only the parameters of the adapter are updated, while the pre-trained parameters remain fixed. This method allows the fine-tuned model to retain knowledge from the pre-training phase while efficiently adapting to the specific downstream task. LoRA \cite{hu2021lora} introduces a low-rank bottleneck by decomposing the weight matrix of the pre-trained model into low-rank matrices. During fine-tuning, only these low-rank matrices are updated, significantly reducing computational costs. QLoRA \cite{dettmers2023qlora} further enhances the LoRA approach to ensure that fine-tuning a 4-bit quantized pre-trained LLM with LoRA does not compromise performance compared to 16-bit full fine-tuning. Utilizing these parameter-efficient fine-tuning strategies allows pre-trained foundation models to be quickly adapted to the medical domain. They deliver expert-level performance and impressive generalizability across multimodal medical imaging.
\end{itemize}

\subsection{Current and Potential Applications}

\begin{itemize}[leftmargin=10pt]
    \item \textbf{Image-to-Text Models}: At present, the principal deployment of multimodal models in the medical domain is primarily within Visual Question Answering (VQA) models \cite{li2023blip}. These models enhance diagnostic efficiency by facilitating the effective analysis of medical images such as X-rays, CT scans, and MRI scans. By posing queries to the VQA model regarding features, abnormalities, or regions of interest within an image, medical practitioners can receive pertinent and detailed answers, aiding in the interpretation of complex medical images. This technology has the potential to reduce diagnostic errors, enhance accuracy, and speed up the diagnostic process, ultimately leading to improved patient outcomes. CLIP (Contrastive Language-Image Pretraining) \cite{radford2021learning} is a vision-language model in the general domain developed by OpenAI. It is trained by learning to associate images and their corresponding textual descriptions using contrastive learning. The pre-trained image encoder can be utilized to obtain image embeddings that facilitate tasks such as image classification, and image-text retrieval. Due to its versatility and impressive cross-modal understanding capabilities, CLIP is a robust tool for combining visual and textual information. MedCLIP \cite{wang2022medclip} enhances and extends the general domain CLIP model to the medical domain by decoupling images and texts for multimodal contrastive learning to employ unpaired medical images and texts. In conjunction with the newly released open-source Language Learning Models (LLMs) such as LLaMA \cite{touvron2023llama}, Alpaca \cite{alpaca}, and Vicuna \cite{chiang2023vicuna} which serve as text decoders, MedCLIP \cite{wang2022medclip} can function as an image encoder in VQA models. XrayGPT \cite{xrayGPT}, a medical domain VQA model similar to miniGPT-4 \cite{zhu2023minigpt} in the general domain, automates the analysis of chest radiographs based on provided X-ray images. Specifically, the pre-trained LLM Vicuna \cite{chiang2023vicuna} is first fine-tuned using medical conversation data comprising approximately 100k real dialogues between patients and doctors and 30k radiology conversations to acquire domain-specific knowledge in the medical domain. Subsequently, a high-quality dataset of 217k chest X-ray findings report from two datasets (MIMIC-CXR and OpenI) are used for fine-tuning the XrayGPT \cite{xrayGPT}, which aligns the frozen MedCLIP as medical image encoder and Vicuna as text decoder to generate chest radiographs summarization from given X-ray images.

    \item \textbf{Text-to-Image Models}: While significant progress has been made in the field of text-prompt image analysis in general domains with models such as Stable Diffusion and DALL·E 2, it remains a formidable challenge in the medical domain \cite{kazerouni2022diffusion}. Medical images, characterized by complex features, varying intensities, and region-specific content, demand a heightened level of precision and strict adherence to specific rules and principles in comparison to general images. The advent of multimodal models has opened prospective applications focusing on text-prompt image analysis. Below are three potential applications of Text-to-Image models.

    \item \textbf{Computer-Aided Diagnosis}: Text-prompt image analysis could enhance diagnostic capabilities. A representative example is ChatCAD, which allows prompting text targeting specific medical image analysis tasks in clinical scenarios, such as emergency rooms, it can bolster the efficiency of diagnosis by doctors and radiologists \cite{wang2023chatcad}. For instance, the 'Segment Anything' model could be deployed in the medical domain by utilizing text prompts that describe the targeted abnormality feature to segment the aberrant lesion. With SAM's zero-shot capability, it is highly adaptable for application in medical imaging across different modalities without the need for specific fine-tuning for each. Furthermore, computer-aided image analysis could be employed for second opinions, comparisons, and decision support, thus leading to more accurate and confident diagnoses.

\begin{figure*}[t]
\label{figure4}
\begin{center}
\includegraphics[width=0.9\textwidth]{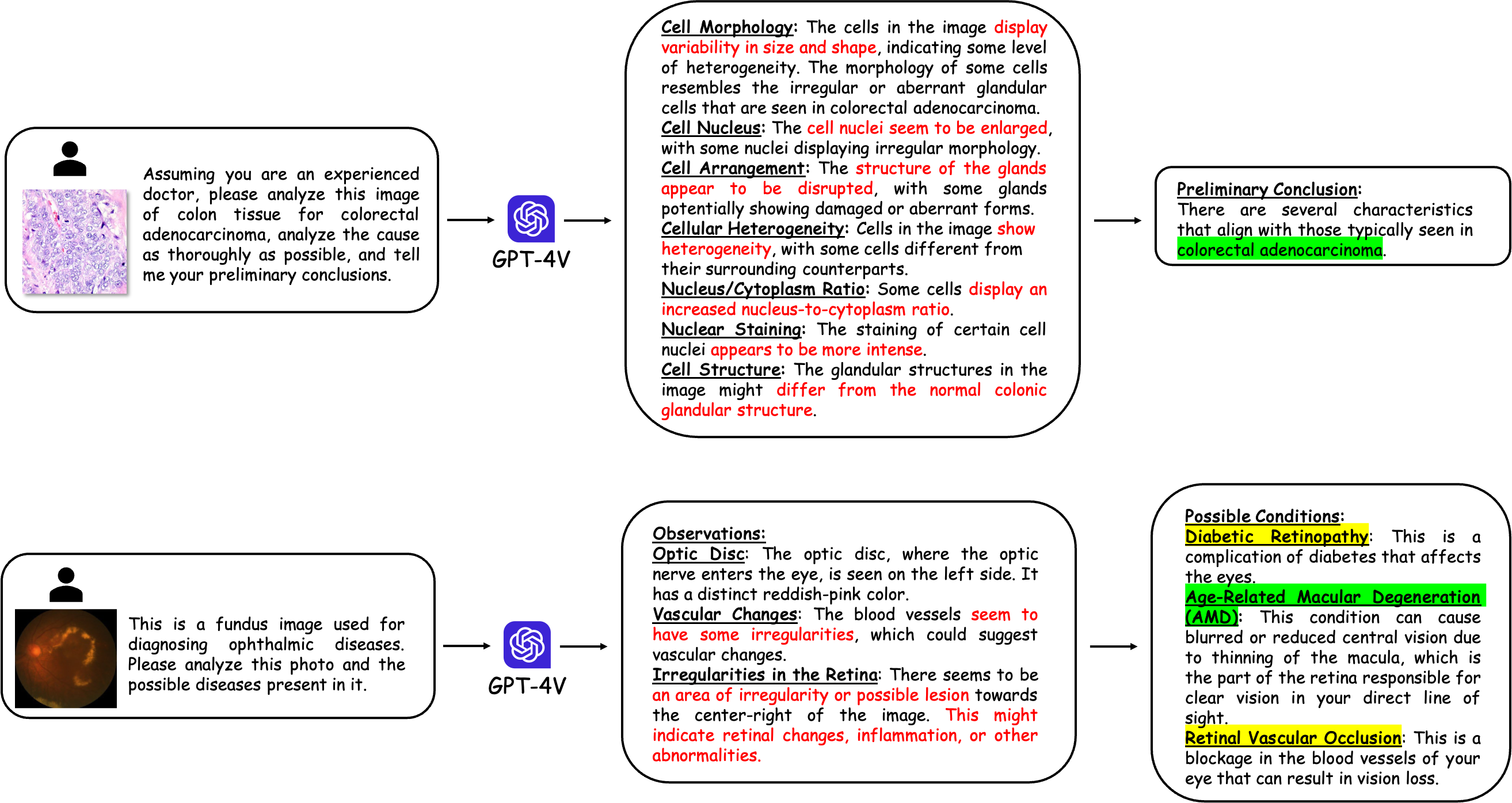}
\end{center}
\caption{An example of using GPT-4V for diagnosing cytological and ocular images, where the red font represents the basis for judgment, green indicates a correct diagnosis, and yellow denotes an uncertain diagnostic outcome.}
\end{figure*}

    
    \item \textbf{Use of Eye-Tracking Technology}: The use of eye-tracking technology can provide valuable data for training LLMs \cite{wang2023chatcad}. By recording where a physician's attention is focused during case study reviews, models like ChatGPT could learn to emulate this prioritization of information, potentially improving the model's diagnostic reasoning capabilities.

    \item \textbf{Use of GPT-4V and Gemini in Medical Imaging}: Accompanying the meteoric advancements in artificial intelligence technologies, multimodal large-scale models have demonstrated profound potential within the realm of medical imaging analysis. The GPT-4V model, introduced by OpenAI in 2023, alongside Google's Gemini model, has particularly captivated the scientific community's interest. Recent study~\cite{liu2023gpt4vtest} indicate that both GPT-4V and Gemini exhibit superior performance in an array of medical disciplines, encompassing neurology, cardiology, cytopathology, and ophthalmology. Fig. 5 delineates instances of application wherein GPT-4V render exceptional analytical prowess in ocular and cytological imaging. GPT-4V, manifests a deep-seated understanding of medical knowledge coupled with the capability to render precise diagnostic verdicts through the synthesis of imaging characteristics. With ongoing advancements, GPT-4V and Gemini are anticipated to soon furnish medical practitioners with formidable adjunctive instruments, bolstering diagnostic precision, streamlining therapeutic approaches, and consequentially ameliorating patient prognoses.

\end{itemize}

\subsection{Challenges and Pitfalls}

\begin{itemize}[leftmargin=10pt]
    \item \textbf{Lack of Paired Data}: One of the most significant challenges in this context is the scarcity of large-scale, paired medical image-text data. This is predominantly due to the heterogeneous and complex nature of medical data. To advance the development and refinement of multimodal models in the medical domain, a large-scale, paired dataset combining visual and textual medical information is essential. A potential solution could be the application of pseudo-labels, such as generated radiology image reports or synthesized medical images. These can serve as synthetic datasets or automatic annotations that may enhance model training and performance. Nevertheless, this approach should be applied cautiously as the quality of synthetic data might fluctuate, and any inaccuracies or inconsistencies could potentially affect the efficacy and robustness of the resulting model.

    \item \textbf{Backdoor Attacks}: The issue of cybersecurity represents another critical concern when deploying large foundational models in healthcare. These models may be vulnerable to backdoor attacks, where malicious actors manipulate the model during training or inference, posing a substantial security threat to downstream users \cite{guan2023badsam}. This flaw could result in a compromised model within a hospital setting, thereby exposing the institution to data leakage and potential security violations. Hence, cybersecurity must be accorded top priority during the deployment and adaptation of multimodal models in the medical domain.

\end{itemize}


\section{Discussion}

Although we have discussed some of the future directions under each section, it is worth providing summarized discussions of the relevant points to have a holistic view of the potential future works. 

\subsection{Potential Bias and Trustworthiness of AGI in Healthcare}
The integration of AGI in healthcare, while promising, raises significant concerns about bias and trustworthiness. AGI models can inadvertently perpetuate or amplify existing biases presented in medical data and practices, potentially leading to disparities in its output, such as diagnosis, treatment plan, etc., across different demographic groups or institutions. These biases may stem from the under-representation of certain populations in training data, historical inequities in healthcare access, or embedded societal prejudices in clinical practice. Ensuring the trustworthiness of AGI in healthcare requires rigorous testing, continuous monitoring, and diverse representation in both development teams and training data \cite{huang2024position}. It's crucial to implement ethical guidelines and regulatory frameworks to govern the use of AGI in medical settings, balancing innovation with equal access and equitable care. As AGI systems become more prevalent in healthcare, maintaining human oversight and fostering a collaborative approach between AI and healthcare professionals will be essential to mitigate these potential issues and build trust among patients and practitioners alike.

\subsection{Facilitating Collaboration across Institutions for AGI Model Development in Medical Image Analysis}
A collaborative approach can address several key challenges in AGI, including data scarcity, model robustness, bias, and generalizability. Training AGI models with pooled data resources and medical expertise can lead to the representation of a wider range of patient populations, imaging equipment, and clinical scenarios. This diversity is essential for AGI models to perform consistently across different healthcare settings and time spans. To overcome the institutional barrier due to data privacy concerns, regulatory compliance across different jurisdictions, and the need to protect intellectual property, it is vital for the field to establish clear governance frameworks, data use agreements, and ethical guidelines to ensure trust and smooth cooperation. Technically, one solution for the collaborative development of AGI is through Federated learning \cite{bonawitz2017practical}, which is a privacy-preserving scheme that addresses data privacy, ownership, and scalability challenges by enabling the training of models across decentralized devices or data sources. In the medical domain, federated learning has demonstrated great promise as it allows the development of powerful and accurate machine learning models while maintaining the privacy and security of sensitive healthcare data \cite{rieke2020future,dayan2021federated}. The advancement of transformer-based AGI models has provided a common foundation for federated learning, enabling different sites to share a standardized architecture. Furthermore, the tokenization process in the transformer architecture contributes to the effectiveness of federated learning. By adopting a standardized token space, models from different centers can maintain a shared understanding of the input data and allow effective data sharing by conducting information exchange at the token level. In summary, an effective combination of AGI and federated learning holds significant promise in developing more powerful systems for medical image analysis.

\subsection{Further Improvement in Expert-Machine Alignment to Learn from Healthcare Expertise}
As previously discussed, achieving expert-machine alignment is much more challenging than the human-machine alignment works done so far, which led to the success of LLMs. To further leverage healthcare expertise in AGI development, we envision that several strategies could be employed. The most straightforward and validated approach is to enhance expert involvement via reinforcement learning and/or model preference optimization \cite{rafailov2024direct}, where clinicians provide feedback on model outputs during training. Active learning sessions \cite{monarch2021human} can also be employed, where clinicians guide AI models through clinical scenarios. In addition, a multidimensional, comprehensive scoring approach would be needed to evaluate model performance across multiple clinical factors, which is a common task in healthcare. Beyond model design, the formation of interdisciplinary teams comprising AI specialists, healthcare professionals, ethicists, and legal experts will be vital for successful AGI development. The expert-machine alignment will facilitate the AGI to learn from clinical expertise, improving its accuracy, ethical compliance, and relevance in practical settings, ultimately leading to a more robust and clinically valuable AI system for medical imaging.

\subsection{Truly Multimodal AGI}
There has been substantial progress in multi-modal generalist foundation models for medical images, such as BiomedGPT \cite{zhang2024generalist}, which could be a good start for a unified approach to processing data from highly heterogeneous sources. We anticipate that multimodal models for medical imaging shall be capable of integrating various data types such as text, images, and tabular (EHR) to enhance medical understanding and reasoning. We envision that LLM will be the reasoning core to improve and coordinate the comprehension, analysis, and inference from multiple data modalities, such as aligning information from medical images and reports. Another important insight from us on the fundamental multi-modal integration is the capability of performing cross-modality data generation. As demonstrated in the works of \cite{chen2024fine}, multimodal AGI can establish a feedback loop between radiology image and text reports via text-guided image generation and image-guided text generation. This bidirectional, cyclic generative scheme is crucial for translating insights from diverse data modalities into actionable clinical recommendations.

\subsection{Safeguarding AGI}
Given the array of tremendous power and the associated potential risks of AGI models, there is a need for comprehensive mitigation strategies to ensure their safe and effective use, especially in the medical domain. As stated in the EU Artificial Intelligence Act, certain AI systems can be considered as "high risk" if they are intended to be used as a safety component of a product or if AI is a stand-alone product. To ensure the responsible use of AGI, several safeguarding measures are essential. These include implementing robust supervision and validation frameworks with interdisciplinary teams and establishing two-way communication for clarification in AGI use, as also delineated in WHO guidance for ethics and governance of artificial intelligence for health \cite{world2021ethics}. The supervision must come with regular audits for risk detection and mitigation as well \cite{chen2019can}. The requirement for "high risk" AI systems from EU Commission also mandates enhanced cyber security, record keeping, and technical documents. In deployment, it is also essential to implement safety guardrails with appropriate disclaimers, plus user education about AGI limitations. Also, because of the nature of certain closed-source, online-based AGI applications such as GPT4, compliance with healthcare data protection laws is imperative. Finally, as discussed in the above expert-machine alignment section, the goal of AGI in medical image analysis is to develop collaborative models that work alongside human doctors, rather than replacing them, so that we can maximize benefits while minimizing risks in clinical practice. The discussion above collectively aim to maintain the integrity of medical practice, protect patient privacy, and ensure that AGI's potential in healthcare is harnessed safely and ethically.

\subsection{AGI Utilization in Clinical Practice: Resource and Speed Concerns}
The deployment of AGI in clinical practice faces significant challenges due to computational constraints and the need for rapid processing \cite{he2019practical}. Healthcare environments often require real-time decision support and swift analysis, particularly in point-of-care and critical care settings, which can make it infeasible to utilize a large AGI model on the premises. Furthermore, healthcare in resource-constrained regions will not be able to access the computational resources needed to run large models. While large foundation models like ChatGPT, GPT-4, and Med-PaLM possess extensive knowledge and robust capabilities, their deployment in these scenarios can be challenging due to computational/network limitations. Thus, we envision that the integration of large-scale and local small-scale models would be highly needed to enhance AGI-assisted medical image analysis across diverse healthcare settings. This leads to the research on alignment between large and small AGI models, which could be achieved via various approaches including Knowledge distillation, transfer learning, and model pruning. Overall, small-scale models can be utilized in time-critical and/or resource-limited scenarios such as emergency rooms, integrated into surgical tools for real-time inference, or employed in home-hospital AI assistants. By bridging the gap between advanced AI capabilities and practical constraints, small-scale models facilitate the accessibility and reach of AGI in medical care.

\section{Conclusion}
The rapid evolution of AGI, exemplified by LLMs and their counterparts in vision and multimodal domains, holds great potential for the field of medical imaging. As our review highlighted, while AGIs possess the disruptive capability to transform various aspects of medical image analysis, from diagnosis to treatment planning, from model development to evaluation criteria, their application in such specialized domains is not without hurdles. Through our exploration, we identified the strategies that guide AGI development and deployment in the medical domain. The present applications, though nascent, offer a glimpse into a future where medical diagnoses, interventions, and research can be significantly augmented by AGI. However, the road ahead requires careful navigation, given the ethical/safety considerations, technical complexities, and practical challenges associated with such integration. As we venture deeper into the AGI era, the future of healthcare appears to be closely tied to the responsible development and application of AI technologies. Despite the challenges, the opportunities offered by AGI are vast. Our hope is that this review not only provides a comprehensive understanding of the current landscape but also inspires future endeavors aimed at harnessing the full potential of AGI in medical imaging, healthcare, and beyond.
\section{Author Contributions}
Author Contributions:
Xiang Li, Pingkun Yan, Wei Liu, Tianming Liu, and Dinggang Shen developed the idea to perform a review of AGI for medical imaging analysis based on a joint panel discussion.
Xiang Li, Lin Zhao, and Lu Zhang wrote the main sections of the manuscript.
Zihao Wu, Zhengliang Liu, Hanqi Jiang, Chao Cao, Shaochen Xu, Yiwei Li, and Haixing Dai performed searching and summarizing of the publications related to each section (LLM, LVM, and multi-modal foundation models).
Yixuan Yuan, Jun Liu, Gang Li, Dajiang Zhu, and Quanzheng Li provided guidance on the review study design and manuscript structuring, as well as necessary resources to implement the review. 
All authors provided critical feedback and helped shape the structure of the review and manuscript writing.



\printbibliography

\end{document}